\documentclass[a4paper]{article}
\usepackage{ICNLSSP2017,amssymb,amsmath,epsfig}
\usepackage{multirow}
\ninept
\def\reg{{\rm\ooalign{\hfil
     \raise.07ex\hbox{\scriptsize R}\hfil\crcr\mathhexbox20D}}}

\title{Sentence Boundary Detection for French with Subword-Level Information Vectors and Convolutional Neural Networks}

\makeatletter
\def\name#1{\gdef\@name{#1\\}}
\makeatother \name{{\em Carlos-Emiliano Gonz\'alez-Gallardo$^1$, Juan-Manuel Torres-Moreno$^1\,^2$}}

\address{$^1$LIA, Universit\'e d'Avignon et des Pays de Vaucluse \\
  $^2$\'Ecole Polytechnique de Montr\'eal \\
{\small \tt carlos-emiliano.gonzalez-gallardo@alumni.univ-avignon.fr, juan-manuel.torres@univ-avignon.fr}}

\begin{document}
\maketitle
\begin{abstract}
In this work we tackle the problem of sentence boundary detection applied to French as a binary classification task ("sentence boundary" or "not sentence boundary"). We combine convolutional neural networks with subword-level information vectors, which are word embedding representations learned from Wikipedia that take advantage of the words morphology; so each word is represented as a bag of their character n-grams.

We decide to use a big written dataset (French Gigaword) instead of standard size transcriptions to train and evaluate the proposed architectures with the intention of using the trained models in posterior real life ASR transcriptions. 

Three different architectures are tested showing similar results; general accuracy for all models overpasses $0.96$. All three models have good F1 scores reaching values over $0.97$ regarding the "not sentence boundary" class. However, the "sentence boundary" class reflects lower scores decreasing the F1 metric to $0.778$ for one of the models.

Using subword-level information vectors seem to be very effective leading to conclude that the morphology of words encoded in the embeddings representations behave like pixels in an image making feasible the use of convolutional neural network architectures.

\end{abstract}

\section{Introduction}

Multimedia resources provide nowadays a big amount of information that automatic speech recognition (ASR) systems are capable to transcribe in a very feasible manner. Modern ASR systems like the ones described in  \cite{fohr2017new} and \cite{le2017disentangling} obtain very low Word Error Rate (WER) for different French sources ($17.10\%$ and $12.50\%$ respectively), leading to very accurate transcriptions that could be used in further natural language processing (NLP) tasks.  

Some NLP tasks like part-of-speech tagging, automatic text summarization, machine translation, question answering and semantic parsing are useful to process, analyze and extract important information from ASR transcriptions in an automatic way \cite{che2016punctuation,lu2010better}. For this to be accomplished a minimal syntactic structure is required but ASR transcriptions don't carry syntactic structure and sentences boundaries in ASR transcriptions are inexistent.

Sentence Boundary Detection (SBD), also called punctuation prediction, aims to restore or predict the punctuation in transcripts. State of the art show that research has been done for different languages like Arabic, German, Estonian, Portuguese and French \cite{tilk2016bidirectional,peitz2014better,zribi2016sentence,kolavr2012development,batista2012bilingual}; nevertheless English is the most common one \cite{che2016punctuation,lu2010better,tilk2016bidirectional,nicola2013improved,ueffing2013improved}. In this paper we focus on French, nevertheless, the proposed architectures and the concepts behind can be used to other languages.

There exist two different types of features in SBD and the use of each type depends of their availability and the methods that will be used. Acoustic features rely on the audio signal and the possible information that could be extracted like pauses, word duration, pitch and energy information \cite{kolavr2012development,igras2016detection,che2016sentence}. Lexical features by contrast, depend on transcriptions made manually or by ASR systems, dealing to textual features like bag of words, word n-grams and word embeddings \cite{che2016punctuation,lu2010better,peitz2014better,zribi2016sentence}.

Conditional random fields classifiers have been used in \cite{lu2010better,nicola2013improved} to predict different punctuation marks like comma, period, question and exclamation marks.
In \cite{kolavr2012development}, adaptive boosting was used to combine many weak learning algorithms to produce an accurate classifier also for period, comma and question marks.
Hand-made contextual rules and partial decision tree algorithms where considered in \cite{zribi2016sentence} to find sentence boundaries in Tunisian Arabic.
In \cite{peitz2014better}, a hierarchical phrase-based translation approach was implemented to treat the sentence boundary detection task as a translation one. 

Deep neural networks were used with word embeddings in \cite{che2016punctuation} to predict commas, periods and questions marks. Three different models were presented: the first one considered a standard fully connected deep neural network while the other two implemented convolutional neural network architectures. Concerning the word embeddings, 50-dimensional pre-trained GloVe word vectors were chosen to perform experiments; this embeddings use a distinct vector representation for each word ignoring the morphology of words. 
Che \textit{et al.} recovered the standard fully connected deep neural network architecture presented in \cite{che2016punctuation}, then an acoustic model was introduced in a 2-stage joint decision scheme to predict the sentence boundary positions.

Following the scheme described in \cite{che2016punctuation}, we aboard the SBD as a binary classification task. The objective is to predict the associated label of a word $w_i$ inside a context window of $m$ words using only lexical features. Audio sources are normally used to train and test SBD models which are not reused for later applications. We want to create models that can be reutilized in further SBD work, so we approach the topic in a different manner using a big written dataset. 

\section{Model Description}

\subsection{Subword-Level Information Vectors}
Subword-Level Information (SLI) vectors \cite{bojanowski2016enriching} are word embedding representations based on the continuos skip-gram model proposed in \cite{mikolov2013distributed} and created using the fastText library\footnote{https://github.com/facebookresearch/fastText}.

Compared to other word embedding representations that assign a distinct vector to each word ignoring their morphology \cite{mikolov2013distributed,pennington2014glove,mnih2013learning,levy2014linguistic}, SLI vectors learn representation for character n-grams and represent words as the sum of those vectors. This provides a major advantage because it makes possible to build vectors for unknown words. Nevertheless for our research we found useful the intrinsic relation between vector's components.

For the present research we used the French pre-trained SLI vectors in dimension 300 trained on Wikipedia using fastText\footnote{https://github.com/facebookresearch/fastText/blob/master/pretrained-vectors.md}.

\subsection{Convolutional Neural Network Models}

Convolutional Neural Networks (CNN) are a type of Deep Neural Network (DNN) in which certain hidden layers behave like filters that share their parameters across space. 

The most straightforward application for CNN is image processing, showing outstanding results \cite{krizhevsky2012imagenet,maggiori2017convolutional}. However they are useful for a variety of NLP tasks like sentiment analysis and question classification \cite{kim2014convolutional}; part-of-speech and named entity tagging, semantic similarity and chunking \cite{collobert2008unified};sentence boundary detection \cite{che2016punctuation} and word recognition \cite{palaz2016jointly} between others.

The input layer of a CNN is represented by a $m$ x $n$ matrix where each cell $c_{ij}$ may correspond to an image's pixel in image processing. For our purpose this matrix represents the relation between a window of $m$ words and their corresponding $n$ dimensional SLI vectors. The hidden layers inside CNN consist of an arrange of convolutional, pooling and fully connected layers blocks.

\subsubsection{Text matrix representation}

Given the intrinsic relation between the components of SLI vectors, we think it is feasible to make an extrapolation to the existing relation between adjacent pixels of an image. This way the $m$ x $n$ matrix of the input layer will be formed by the context window in (\ref{eq:window}) where $w_i$ is the word for which we want to get the prediction. The columns of the matrix will be represented by each one of $n$ components of their corresponding SLI vectors.

\begin{equation}
	\{w_{i-(m-1)/2},...,w_{i-1},w_i,w_{i+1},...,,w_{i+(m-1)/2}\}
    \label{eq:window}
\end{equation}

\subsubsection{CNN-A}

The hidden architecture of the first model (Figure \ref{fig:models} (CNN-A)) is based on a model presented in \cite{che2016punctuation}. It is composed of three convolutional layers (A\_conv-1, A\_conv-2 and A\_conv-3), all three with valid padding and stride value of one. A\_conv-1 has a 2x4-shape kernel and 64 output filters, A\_conv-2 has a 2-shape kernel and 128 output filters and A\_conv-3 has a 1x49-shape kernel and 128 output filters. After A\_conv-1, a max pooling layer (A\_max\_pool) with 2x3-shape kernel and stride of 2x3 is staked. After the convolution phase, two fully connected layers (A\_f\_c-1 and A\_f\_c-2) with 4096 and 2048 neurons each and a final dropout layer (A\_dropout) are added. The output of all convolutional, max pooling and fully connected layers are in function of RELU activations.

\subsubsection{CNN-B}
	
In our second model (Figure \ref{fig:models} (CNN-B)) we tried to reduce the complexity generated by the three convolution layers of CNN-A. For this model there are only two convolutional layers (B\_conv-1 and B\_conv-2), both with valid padding and stride value of one. B\_conv-1 has a 3-shape kernel and 32 output filters while B\_conv-2 has a 2-shape kernel and 64  output filters. To downsample and centralize the attention of the CNN in the middle word of the window, a max pooling layer (B\_max\_pool) with 2x3-shape kernel and stride of 1x3 is implemented after B\_conv-2. The final part of the CNN is formed by 3 fully connected layers (B\_f\_c-1, B\_f\_c-2 and B\_f\_c-3) with 2048, 4096 and 2048 neurons each and a dropout layer (B\_dropout) attached to B\_f\_c-3. The output of all convolutional max pooling and fully connected layers are in function of RELU activations.

\subsubsection{CNN-C}
	
\begin{figure}[h]
	\includegraphics[trim={0 0cm 0 0cm},width=1\columnwidth]{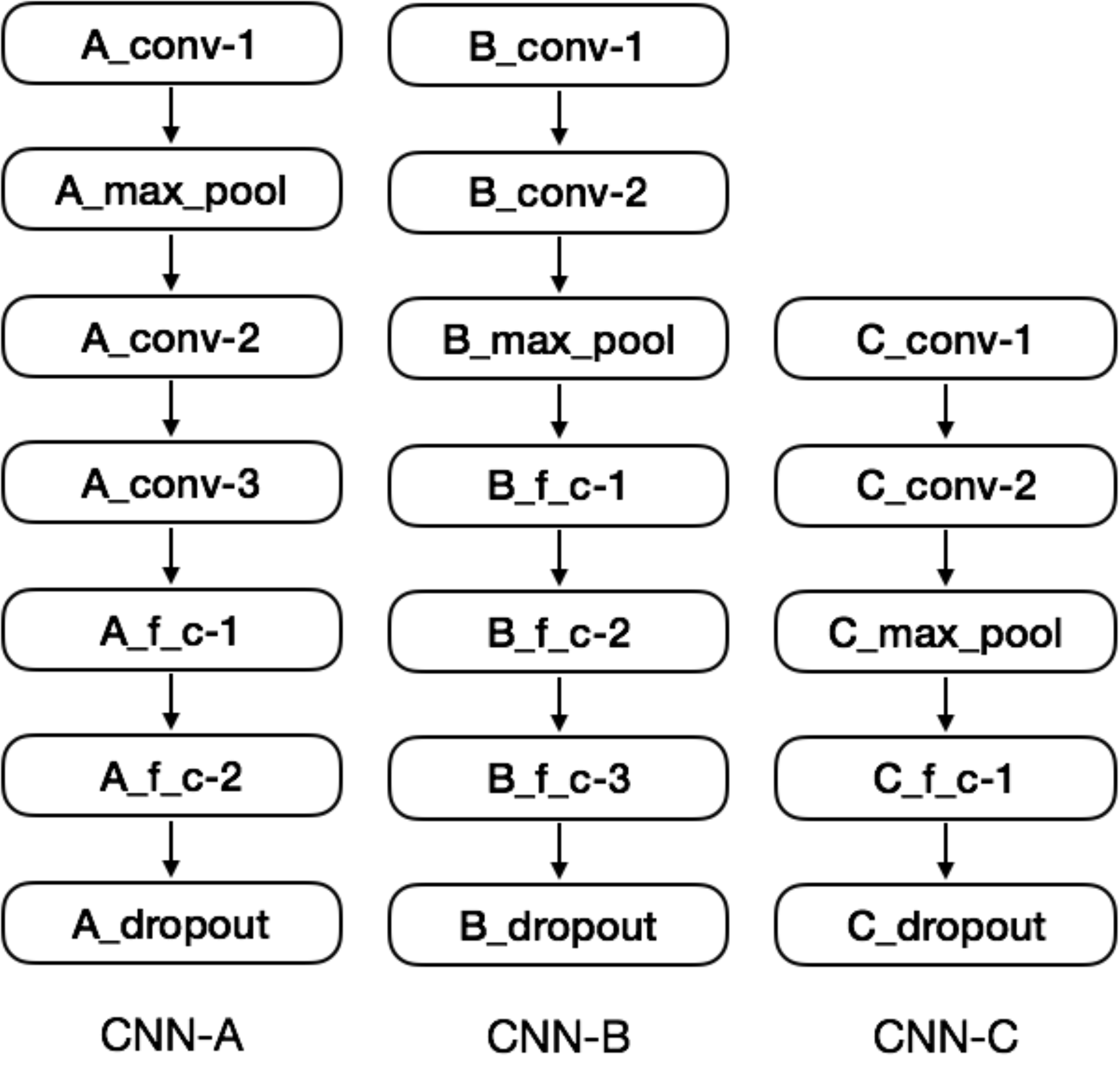}	
	\caption{CNN hidden architectures}
	\label{fig:models}
\end{figure}

Finally, in our third model (Figure \ref{fig:models} (CNN-C)) we simplified the fully connected layers of CNN-B. The convolutional and max pool layers (C\_conv-1, C\_conv-2 and C\_max\_pool) are the same than in CNN-B. For this model, only one fully connected layer of 2048 neurons is present (C\_f\_c-1) which is attached to a dropout layer (C\_dropout).The output of all convolutional max pooling and fully connected layers are in function of RELU activations.

\section{Experimental Evaluation}
\subsection{Dataset}

SBD experimentation datasets normally rely on automatic or manual transcriptions to train and test the proposed systems \cite{che2016punctuation,peitz2014better,ueffing2013improved}. As shown in Table \ref{tbl:corpusstats}, the amount of tokens is, in average 21.25k, which only 2.5k (12.48\%) correspond to any punctuation mark.

\begin{table} [th]
\caption{\label{tbl:corpusstats} {\it Oral datasets.}}
\vspace{2mm}
\centerline{
\begin{tabular}{|c|c|c|c|}
\hline
Dataset & tokens & punctuation & percentage \\
\hline 
\cite{nicola2013improved} WSJ & 51k & 5k & 9.8\% \\
\cite{nicola2013improved} TED Ref & 17k & 2k & 11.8\% \\
\cite{nicola2013improved} TED ASR  & 17k & 2k & 11.8\% \\
\cite{nicola2013improved} Dict & 25k & 3k & 12\% \\
\cite{che2016punctuation} Ref & 13k & 2k & 15.4\% \\
\cite{che2016punctuation} ASR & 13k & 2k & 15.4\% \\
\textbf{Average} & \textbf{21.25k} & \textbf{2.5k} & \textbf{12.48\%} \\
\hline
\end{tabular}}
\end{table}

In order to reuse the proposed architectures and trained models for real life ASR transcriptions and further NLP applications we opted for a big written dataset. It consists of one section of the French Gigaword First Edition\footnote{https://catalog.ldc.upenn.edu/LDC2006T17} (GW\_afp) created by the Linguistic Data Consortium. Before any experimentation, the following normalization rules were applied during a preprocessing cleaning process over the GW\_afp dataset:

\begin{itemize}
\item XML tags and hyphens elimination
\item Lowercase conversion
\item Doubled punctuation marks elimination
\item Apostrophes isolation
\item Substitution of (?, !, ;. :, .) into "$<SEG>$"
\end{itemize}

The amount of tokens after the cleaning process for the GW\_afp dataset is 477M, where $9\%$ correspond to any punctuation mark (Table \ref{tb2:corpusGW}). This proportion is very similar to the \textit{Nicola et al. (2013) WSJ}'s dataset presented in Table \ref{tbl:corpusstats}, which consists of newspaper text. $80\%$ of the tokens were used during training and validation while $20\%$ was used exclusively for testing.

\begin{table} [th]
\caption{\label{tb2:corpusGW} {\it GW\_afp dataset statistics.}}
\vspace{2mm}
\centerline{
\begin{tabular}{|c|c|c|c|}
\hline
Dataset & tokens & punctuation & percentage \\
\hline 
GW\_afp & 477M & 43M & 9\% \\
\hline
\end{tabular}}
\end{table}

\subsection{Metrics}

To evaluate our models we considered necessary two types of metrics. At a first glance we opted for Accuracy (\ref{eq:accuracy}), a general metric that could measure the performance of the models regardless the class. Nevertheless, given the disparity of samples between the two classes, Accuracy is very likely to be biased; so Precision (\ref{eq:precision}), Recall (\ref{eq:recall}) and F1 (\ref{eq:fscore}) metrics were calculated for each one  of the two classes.

\begin{equation}
	Accuracy = \frac{\# correctly\ predicted\ samples}{\# samples}
    \label{eq:accuracy}
\end{equation}

\begin{equation}
	Precision_{ci} = \frac{\# correctly\_predicted\_samples_{ci}}{\# total\_predicted\_samples_{ci}}
    \label{eq:precision}
\end{equation}

\begin{equation}
	Recall_{ci} = \frac{\# correctly\_predicted\_samples_{ci}}{\# total\_samples_{ci}}
    \label{eq:recall}
\end{equation}

\begin{equation}
	F1_{ci} = 2*\frac{Precision_{ci}*Recall_{ci}}{Precision_{ci}+Recall_{ci}}
    \label{eq:fscore}
\end{equation}

\subsection{Results}

\begin{table*} [th]
\caption{\label{tb3:results} {\it Results for CNN models.}}
\vspace{2mm}
\centerline{
\begin{tabular}{|c|c|c|c|c|c|c|c|}
\hline
Model & Accuracy & \multicolumn{2}{c|}{Precision} & \multicolumn{2}{c}{Recall} & \multicolumn{2}{|c|}{F1}\\
&&NO\_SEG&SEG&NO\_SEG&SEG&NO\_SEG&SEG\\
\hline 
CNN-2 \cite{che2016punctuation} & - & - &0.836& - &0.723 & - & 0.775\\
CNN-2A \cite{che2016punctuation} & - & - &0.776& - &\textbf{0.799} & - & 0.788\\
CNN-A\_u & 0.909 & 0.909 &0& 1 &0 & 0.952 & 0\\
\hline 
CNN-A & 0.963& 0.972 &\textbf{0.853}& \textbf{0.988} &0.718 & 0.980 & 0.778\\
CNN-B & \textbf{0.965}& \textbf{0.975} &0.845& 0.986 &0.754 &\textbf{0.981} & \textbf{0.795}\\
CNN-C & 0.963& 0.974 &0.832& 0.985 &0.75 & 0.980 & 0.787\\
\hline
\end{tabular}}
\end{table*}

Three different baselines are shown in Table \ref{tb3:results}. In their experiments, Authors of \cite{che2016punctuation} compute only Precision, Recall and F1 for the "sentence boundary" class. CNN-2 and CNN-2A refer to the same convolutional neural network model but in CNN-2A is only taken into account half the value of softmax output for the "no sentence boundary" class. This variation equilibrates Precision and Recall of CNN-2 reaching a F1 score value of  $0.788$.

CNN-A\_u refers to the untrained CNN-A model. We wanted to have this as a baseline to visualize how the unbalanced nature of the samples impacts all measures and may mislead general metrics like Accuracy.

Accuracy over all the proposed models is higher than in CCN-A\_u, reaching the higher score for CNN-B. Concerning Precision, CNN-B and CNN-A overperform for different classes. CNN-2A reflects a higher Recall than the rest of the baselines and models. Finally, F1 score for both classes is higher in CNN-B.

\begin{figure}[h]
	\includegraphics[trim={0 0cm 0 0cm},width=1\columnwidth]{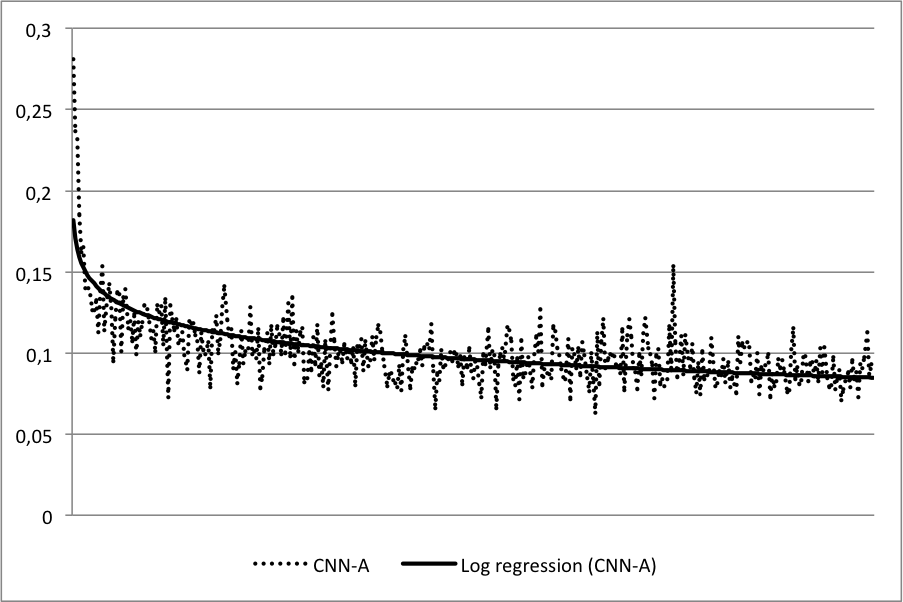}	
	\caption{Cross entropy (CNN-A)}
	\label{fig:ccna_loss}
\end{figure}

\begin{figure}[h]
	\includegraphics[trim={0 0cm 0 0cm},width=1\columnwidth]{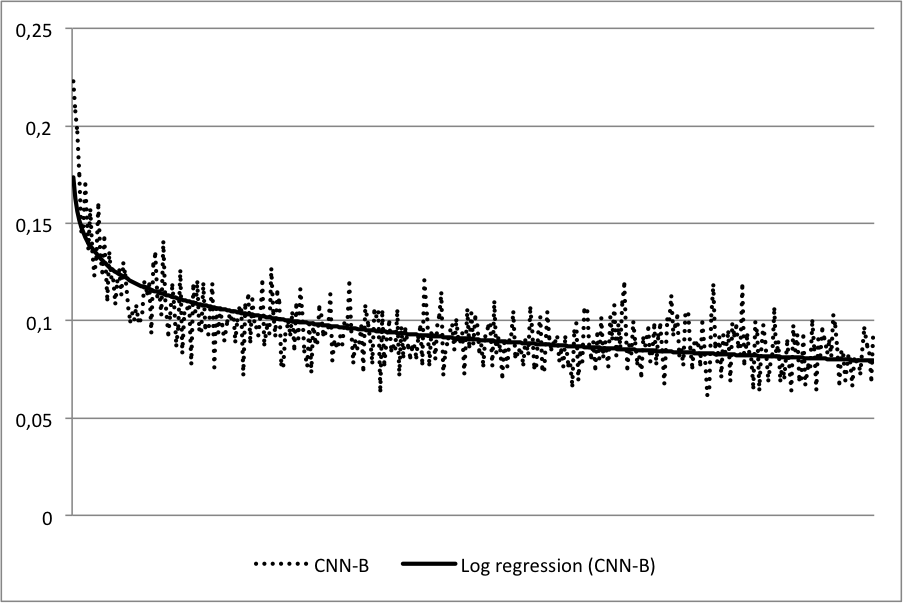}	
	\caption{Cross entropy (CNN-B)}
	\label{fig:ccnb_loss}
\end{figure}

\begin{figure}[h]
	\includegraphics[trim={0 0cm 0 0cm},width=1\columnwidth]{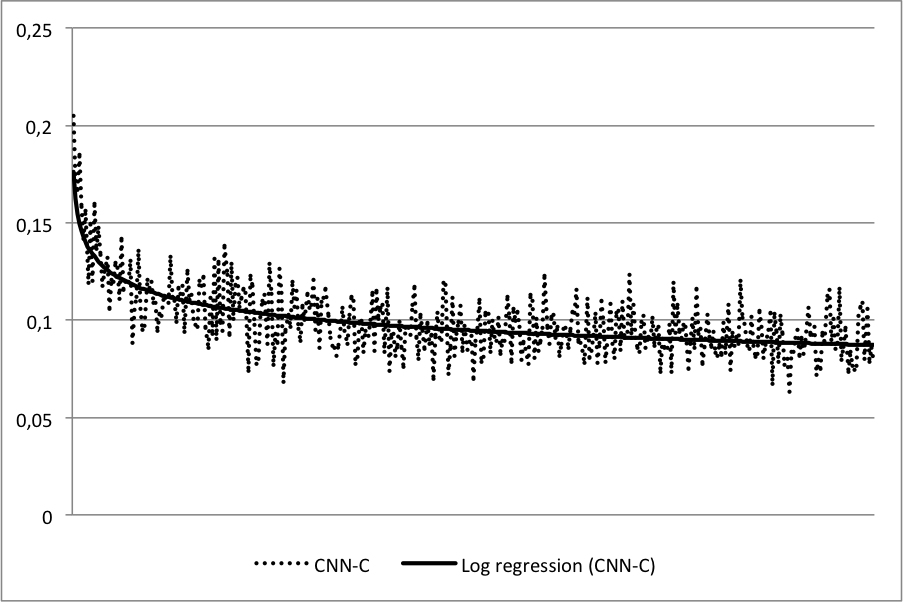}	
	\caption{Cross entropy (CNN-C)}
	\label{fig:ccnc_loss}
\end{figure}

Given the similar results of the models we wanted to see the behavior of the models during training process. Cross entropy during training process is plotted in Figures \ref{fig:ccna_loss} to \ref{fig:ccnc_loss}. The three curves show a similar behavior and converge in a value below $0.09$. CNN-B slightly overperforms the rest of the models (Table \ref{tb3:cross}).

\begin{table} [th]
\caption{\label{tb3:cross} {\it Cross entropy during training}}
\vspace{2mm}
\centerline{
\begin{tabular}{|c|c|}
\hline
Model & Cross entropy \\
\hline 
CNN-A & 0.082 \\
CNN-B & \textbf{0.080} \\
CNN-C & 0.089 \\
\hline
\end{tabular}}
\end{table}

\section{Conclusions}

In this paper we combined CNN networks with SLI vectors to tackle the problem of sentences boundary detection as a binary classification task for French. We used a big written dataset instead of standard size transcriptions to reuse the trained models in further transcriptions.
SLI vectors, that represent words as the sum of their characters vectors taking advantage of their morphology, showed to be very effective working with our three CNN models.
In a future, we will include other languages like Arabic and English. Also we will reuse the trained models in a variety of ASR transcriptions of newscasts and reports domain.

\section{Acknowledgements}
We would like to acknowledge the support of Chist-Era for funding this work through the  \textit{Access Multilingual Information opinionS (AMIS)}, (France - Europe) project.

\newpage
\eightpt
\bibliographystyle{IEEEtran}
\bibliography{refs}
\end{document}